\documentclass[11pt,a4paper]{article}
\usepackage[hyperref]{acl2021}
\usepackage{times}
\usepackage{latexsym}

\usepackage{graphicx}
\usepackage{amsmath}
\usepackage{amsfonts}
\usepackage{multicol}
\usepackage{xcolor}
\usepackage[utf8]{inputenc}
\usepackage{booktabs}
\usepackage{verbatim}
\usepackage{array}
\usepackage{dblfloatfix}

\usepackage{comment}
\usepackage{url}
\usepackage{multirow}
\usepackage{multicol}
\usepackage{booktabs}
\usepackage{colortbl}
\usepackage{caption}
\usepackage{rotating}

\definecolor{ao(english)}{rgb}{0.0, 0.5, 0.0}

\usepackage{microtype}

\aclfinalcopy 

\title{\textsc{CO-NNECT}: A Framework for Revealing Commonsense Knowledge Paths as Explicitations  of Implicit Knowledge in Texts}

\author{Maria Becker, Katharina Korfhage, Debjit Paul, Anette Frank \\
  Department of Computational Linguistics, Heidelberg University\\
  \texttt{mbecker|korfhage|paul|frank@cl.uni-heidelberg.de} \\}

\date{}

\begin{document}
\maketitle
\begin{abstract}
In this work we leverage commonsense knowledge in the form of knowledge paths to establish
connections between sentences, as a form of \textit{explicitation of implicit knowledge}. Such connections can be direct (singlehop paths) 
or require intermediate concepts (multihop paths). 
To
construct
such paths 
we combine two model types in a joint framework
we call \textsc{Co-Nnect}: a \textit{relation classifier}  
that predicts
direct connections between 
concepts; and a \textit{target prediction model} that generates target or intermediate concepts given a source concept and a relation, 
which we use to construct
multihop paths. 
Unlike
prior work that relies exclusively on \textit{static} knowledge 
sources, we leverage 
language models
finetuned on knowledge 
stored in 
ConceptNet, to 
\textit{dynamically} generate knowledge paths, as explanations of 
implicit knowledge that connects sentences in 
texts. 
As a central contribution 
we design manual and automatic evaluation settings for assessing the \textit{quality} of the generated paths.
We conduct evaluations
on two argumentative datasets and show that a combination of the two 
model types generates meaningful, high-quality knowledge paths between
sentences 
that
reveal implicit knowledge conveyed in
text.
\end{abstract}

\section{Introduction}
\label{sec:intro}

Commonsense knowledge covers simple facts 
about the world, people and  everyday life, e.g., \textit{Birds can fly} or \textit{Cars are used for driving}. 
It is increasily used for many NLP 
tasks, 
e.g. for question answering \cite{OpenBookQA2018}, 
textual entailment \cite{weissenborn2018}, 
or classifying argumentative functions \cite{pauletal:2020}. 
In this work, we leverage commonsense knowledge in the form of single- 
\textit{and} multihop 
knowledge paths for establishing connections between concepts from different sentences in texts, and show that these 
paths can 
\textit{explicate} implicit information conveyed by the text. Connections can either be direct, e.g. given the sentences \textit{The car was too old} and \textit{The engine broke down}, the concepts \textit{car} and \textit{engine} can be linked with a direct relation (singlehop path) \textit{car $\rightarrow$ \textsc{HasA} $\rightarrow$ engine}; or indirect -- here intermediate concepts are required to establish the link, as 
between 
\textit{Berliners produce too much waste} and \textit{Environmental protection should play a more important role}, where the link between \textit{waste} and \textit{environmental protection} requires a multihop reasoning path:
\textit{waste $\rightarrow$ \textsc{ReceivesAction} $\rightarrow$ recycle $\rightarrow$ \textsc{PartOf} $\rightarrow$ environmental protection}. 

We show that two complementary model types 
can be combined
to 
solve 
the 
two subtasks:
(i) for predicting singlehop paths 
between concepts, we propose a \textit{relation classification} model that is very precise, 
but restricted to 
direct connections between concepts;
(ii) for constructing longer paths we rely on a \textit{target prediction} model that can  generate intermediate concepts and is thus able to generate
multihop paths. 
However, the intermediate concepts 
can be
irrelevant or misleading.
To our knowledge, 
prior work has applied \textit{either} relation classification \textit{or}
target prediction models. 
We propose \textsc{Co-Nnect}, a framework 
that 
establishes \textbf{\textsc{Co}}mmonsense knowledge paths for \textbf{\textsc{Connect}}ing sentences 
by \textit{combining} relation classification and target prediction models, leveraging their strengths and minimizing their weaknesses. 
With  \textsc{Co-Nnect}, we
obtain \textit{
high-quality knowledge paths} that explicate implicit knowledge
conveyed by the text.

We focus on commonsense knowledge 
in ConceptNet \cite{speer17}, a knowledge graph (KG) that represents concepts (words or phrases) as
nodes, 
and 
relations 
between them as edges, e.g.,\ $\langle$\textit{oven},\textsc{UsedFor}, \textit{baking}$\rangle$. 
As instances
of the model types we use COREC \cite{beckeretal:2019}, a 
multi-label relation classifier that predicts \textit{relation types} and that we enhance with a pretrained language model; and COMET \cite{comet}, a pretrained transformer model 
that learns to generate \textit{target concepts} given 
a source concept 
and a relation. In contrast to models that retrieve knowledge from static KGs \cite{OpenBookQA2018,Lin2019KagNetKG}, 
both models are fine-tuned on ConceptNet and applied \textit{on the fly}, to dynamically generate knowledge paths that generalize beyond the static knowledge, allowing us to predict unseen knowledge paths.
We 
compare our models to a baseline model that solely 
relies 
on static KGs. 

We evaluate our framework
on two English argumentative datasets, IKAT \cite{beckeretal:2020a} and ARC \cite{habernal-etal-2018-argument}, which offer annotations that explain implicit connections between 
sentences. 
While knowledge paths have been widely used in NLP downstream tasks, a careful evaluation of these paths 
has not received much attention. As a central contribution of our work, we
 address this shortcoming 
by designing manual and automatic settings for path evaluation: 
we evaluate the relevance and quality of the paths and their ability to represent implicit knowledge in an annotation experiment; and we 
compare the 
paths to the annotations of implicit knowledge 
in  IKAT and ARC, using 
automatic similarity metrics.

Our main contributions are: i) we propose \textsc{Co-Nnect}, a framework 
that combines two complementary types of knowledge path prediction models
that 
have previously only been applied separately;\footnote{The code for our framework can be found here: \url{https://github.com/Heidelberg-NLP/CO-NNECT}.} 
ii) we show that commonsense knowledge paths generated with \textsc{Co-Nnect} effectively represent implied knowledge between sentences;
iii) we propose an evaluation scheme that 
measures the quality of the knowledge paths, going beyond
many approaches that use knowledge paths for downstream applications without analyzing their quality.


\section{Related Work}
\label{sec:relatedwork}

In this work we combine relation classification and target prediction for generating
commonsense knowledge representations over text. 
\textbf{Relation classification} covers a range of methods and learning paradigms for representing relations. 
A variety of neural architectures such as RNNs \cite{zhang18}, CNNs \cite{guo-etal-2019-attention}, 
sequence-to-sequence models \cite{trisedya-etal-2019-neural} 
or language models \cite{Wu2019}
achieved state-of-the-art results. 
To our knowledge, 
\citet{beckeretal:2019} 
is the only work that 
proposed a relation classification model 
specifically for ConceptNet relations, which we adapt for our work.
Besides COMET \cite{comet}, the model used in our approach, 
\citet{saito18} perform \textbf{target prediction} on ConceptNet using an attentional encoder-decoder model.
They improve 
the KB completion model of \citet{li-EtAl:2016:P16-14} by jointly 
scoring triples and predicting target concepts. 

\textbf{Utilizing commonsense knowledge paths.} 
When using commonsense knowledge for 
question answering \cite{OpenBookQA2018}, commonsense 
reasoning \cite{Lin2019KagNetKG} or NLI 
\cite{Kapanipathi}, most approaches 
rely on paths retrieved from \textit{static} knowledge resources.
In contrast,
we propose a framework 
that in addition
makes use of \textit{dynamic} knowledge provided by language models.
Few other models have used 
 knowledge paths \textbf{dynamically}, e.g. 
\citet{pauletal:2020}, who enrich ConceptNet on the fly 
when classifying  argumentative functions. \\
\citet{Wang2020ConnectingTD} make use of \textbf{language models} 
for question answering. They
 generate multihop 
paths 
by sampling random walks 
from ConceptNet 
and finetune a language model on these paths to connect question and answers, 
improving accuracy 
on two 
question answering benchmarks. 
\citet{Bosselut2019DynamicKG} 
generate 
knowledge
paths using a
language model 
for zero-shot question answering, which
they use 
to select the 
answer to a question, surpassing performance of pretrained language models 
on SocialIQA (a multiple-choice question answering dataset for probing machine’s emotional and social intelligence in a variety of everyday situations).
Similarly, \citet{chang-etal-2020-incorporating} incorporate knowledge from ConceptNet in pretrained language models for SocialIQA. They extract keywords from question and answers, 
query ConceptNet for relevant triples, and incorporate them in their language models via attention. Their evaluation shows that their knowledge-enhanced 
model outperforms knowledge-agnostic baselines. 
Finally, \citet{paul-frank-2020-social} propose an 
attention model that encodes commonsense inference rules and incorporates them in a transformer based reasoning cell, taking advantage of pretrained language models and structured knowledge. 
Their evaluation on two reasoning tasks 
shows that their model improves performance over 
models 
that lack external knowledge. 
Hence, none of these systems \textit{directly} evaluates the \textbf{quality} of the generated paths, but measure the effectiveness of 
commonsense knowledge \textit{indirectly} by evaluation on
downstream  tasks. 
We will address this shortcoming in our work by carefully evaluating the quality of the generated paths.

\section{Enriching Texts with Commonsense Knowledge Paths
}

\label{sec:approaches}

\begin{figure*}[t]
\begin{centering}
\includegraphics[width=1
\textwidth]{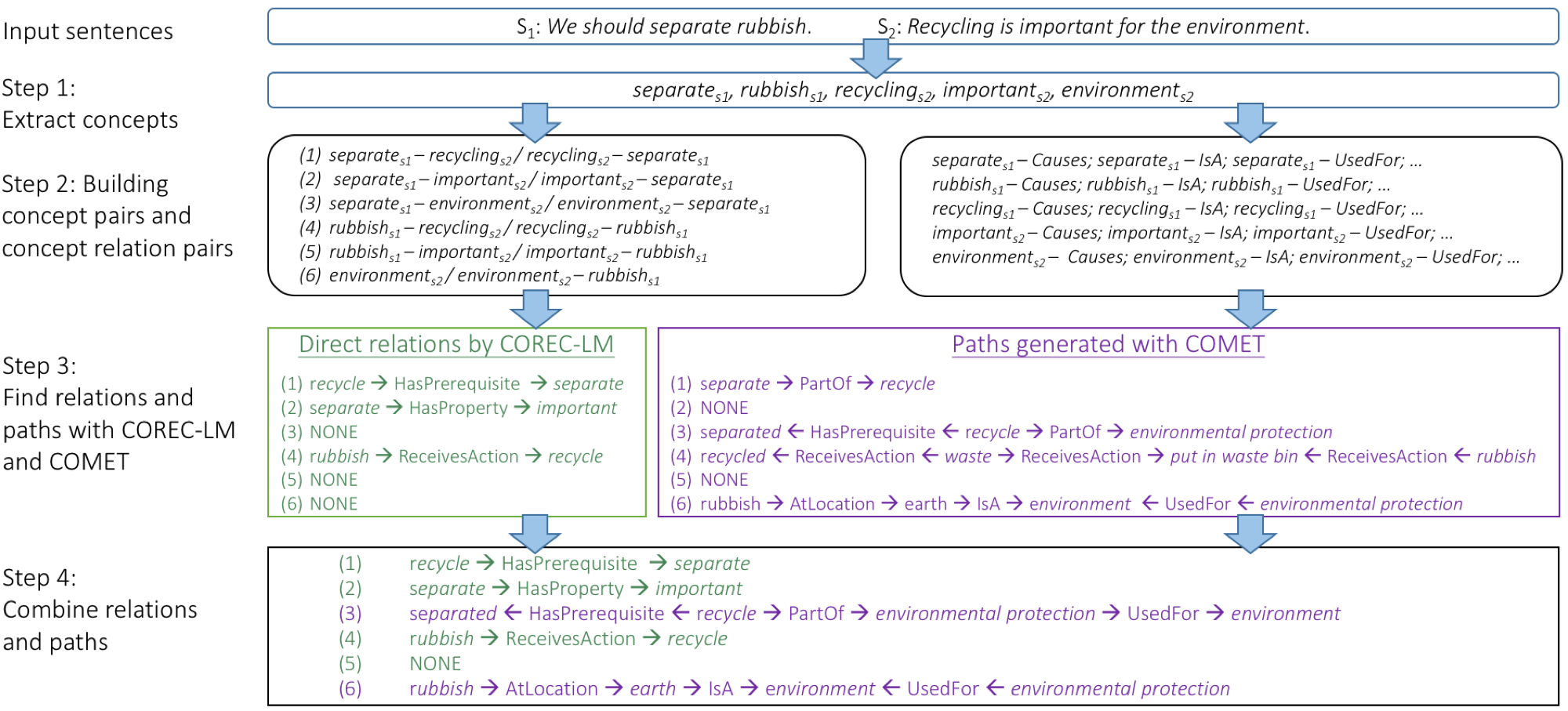}
\caption{Our framework \textsc{Co-Nnect}: It 
finds single- and multihop paths between concepts, as explicitations of implicit knowledge 
that connects  sentences.} 
\label{fig:pipe}
\end{centering}
\end{figure*}

This section describes \textsc{Co-Nnect}, the framework we propose for enriching texts with commonsense knowledge, by establishing relations or paths between concepts from different sentences. Towards this aim, we apply 
relation classification and target prediction models in  combination.
We first characterize
differences between
the two model types and their instantiations, COREC-LM and COMET, describe how we adapt them to our task and 
evaluate them on ConceptNet to assess their performance 
(\S 3.1). We then show how we utilize the models 
to establish connections 
between concepts in texts (\S 3.2) and present a baseline model that uses ConceptNet as a static 
KG to establish commonsense knowledge paths
(\S 3.3).

\subsection{Comparing and Evaluating Model Types} 

Relation classification and target prediction both aim at representing relational commonsense knowledge, but 
the respective task settings are fundamentally different. 
We choose two models that have been developed for representing commonsense knowledge in CN: COREC, a relation classification and COMET, a target prediction model. 

\textbf{Relation Classification with COREC-LM.} 
A relation classifier is ideally suited to predict
\textit{direct} relations between concepts, 
hence we can apply
COREC \cite{beckeretal:2019}, an open-world multi-label relation classification system, for this task. 
Given a pair of concepts 
$c_s,c_t$ from sentences, 
it predicts one or several relations $r_i$ from a set of relation types $R_{\textsc{cn}}$ that hold between $c_s$ and $c_t$. We enhance the original neural model 
with the pretrained language model DistilBERT \cite{Sanh2019DistilBERTAD} 
to construct a classifier we call 
COREC-LM. 
We finetune this model on ConceptNet by masking the relations and use sigmoid as output layer to model the probability of each relation independently, 
accounting for
ambiguous relations in CN.

\textbf{Target Prediction with COMET.} 
To generate multihop paths 
 that include (possibly novel) intermediate concepts, 
we apply COMET \cite{comet}, a transformer encoder-decoder based on GPT-2
\cite{radford2019language}.
Input to the model is 
a source concept $c_s$ and a relation $r_i$. 
Then the pretrained language model 
is finetuned using ConceptNet as labelled train set for the task of generating new concepts. 
Depending on the beam size, COMET can propose multiple targets per input instance. 

\textbf{Datasets.} 
To compare model performances, we evaluate COREC-LM and COMET on the \textbf{CN-100k} benchmark dataset of \citet{li-EtAl:2016:P16-14}, which is based on the OMCS  subpart of ConceptNet. The dataset comprises 37 relation types  such as \textsc{IsA, PartOf} or \textsc{Causes} and contains 100k relation triples in the train set and 1200 in the development and the test set, respectively. CN-100k contains a lot of infrequent relations which are hard to learn and often overspecific (e.g. \textsc{HasFirstSubevent}), and hence not useful for establishing high quality relations and paths between concepts. We therefore extract a subset that contains all triples of the 13 most frequent relations (\textbf{CN-13}).\footnote{These are: \textsc{AtLocation, Causes, CapableOf, IsA, HasPrerequisite, HasProperty, HasSubevent}, \textsc{UsedFor}, \textsc{CausesDesire, Desires, HasA, MotivatedByGoal} and \textsc{ReceivesAction}.} CN-13 covers 90,600 triples for training, 1080 triples for development, and 1080 triples for testing.

Since our 
application task 
requires that the relation classifier also learns to detect that 
a given concept pair is \textit{not} related, we extend the data for training and testing COREC-LM with a \textsc{Random} class that contains unrelated concept pairs, which we add to CN-100k and CN-13.\footnote{For details about the construction of the \textsc{Random} class, cf. Appendix.} 

\textbf{PoS Sequence Filtering.} We apply a type-based PoS sequence filtering for COREC-LM and COMET, where the type is dependent on the predicted relation. The relation \textsc{IsA}, for example, is supposed to connect two noun phrases; in contrast, \textsc{HasPrerequisite} typically relates two verb phrases. We determine frequent PoS sequence patterns for specific argument types from the ConceptNet resource and use them to filter relation and path predictions. 

\textbf{Metrics.} We evaluate COREC-LM in terms of weighted F1-scores, precision and recall, which is its genuine evaluation setting. For COMET we report precision scores for the first prediction with highest confidence score (hits@1); we further report hits@10 which gives information if the correct triple is included in the first ten predictions (which will be important since we later use a beamsize of 10 for generating paths). 
In addition, we report accuracy using the Bilinear AVG model of \citet{li-EtAl:2016:P16-14} (COMET's genuine evaluation setting), which is trained on CN-100k and produces a probability for a generated relation triple to be correct. Following \citet{comet}, we apply a beamsize of 1 and a threshold at 0.5 for judging a triple as correct.

\textbf{Model Performances.} \textbf{COREC-LM} achieves high F1-scores on CN-100k (76.5) and CN-13 (86.0).\footnote{The original version of COREC 
\cite{beckeretal:2019} achieves F1 of 53.31/CN-100k; 72.33/CN-13.}
Scores are significantly lower when adding 
the \textsc{Random} class (-7pp on CN-100k\&CN-13), indicating that detecting unrelated concept pairs 
is not trivial. The results 
show that a strength of COREC-LM 
is its precision (90.1/CN-100k; 88.2/CN-13) -- which we will leverage 
when combining 
models. 
\textbf{COMET} achieves high accuracy scores (92.3/CN-100k; 96.3/CN-13) according to the bi-score. 
For the much stricter metric hits@1 which judges a 
triple as correct only if it matches the respective triple in the testset, much lower scores are achieved (25/CN-100k; 23.5/CN-13), which is evident given the wide range of possible target generations. Higher scores for hits@10 (65.3/CN-100k; 65.9/CN-13) 
show that the chance for correct predictions significantly rises with increasing beam size. 


\textbf{In sum}, COREC-LM and COMET both aim at learning commonsense knowledge representations, but tackle different tasks and have different strengths and weaknesses. 
COREC-LM is very precise in its predictions, but is restricted to predicting  \textit{direct} relations between
two given concepts. COMET is more powerful since it can 
genuinely generate \textit{novel} target concepts and thus can generate multihop paths. However, it tends to be more imprecise, 
and bears the 
risk of generating \textit{irrelevant or noisy} concepts. 
Hence, a combination of models seems beneficial, 
to predict high-quality single- \textit{and} multihop paths between concepts. 

\subsection{Establishing Connections 
Using Relation Classification and Target Prediction}

In the following we describe how we combine and apply COREC-LM and COMET in a joint framework, \textsc{Co-Nnect}, to establish
high-quality knowledge paths between sentences. An overview is given in Fig.\ref{fig:pipe}.
In the first step 
we \textbf{extract relevant concepts} 
from the text. 
For this
we integrate the concept 
extraction tool \textsc{CoCo-Ex} \cite{beckeretal:2021:COCOEX}, 
which extracts meaningful concepts from 
texts and maps them to concept nodes in CN, considering all surface forms. 

\textbf{Linking Concepts with \textit{Direct Relations.}}
We construct
all possible pairs of concepts extracted from $S_1$ and $S_2$ by taking 
the cross product
$c_s$$\times$$c_t$, where $c_s$ is a concept from $S_1$, and $c_t$ a concept from $S_2$ (Fig. \ref{fig:pipe}, Step 2, left). 
We then apply COREC-LM trained
on \textsc{CN-13}+\textsc{Random} with a tuned threshold of 0.9 for predicting which relation $r_i$ $\in$ $R_{CN13}$ holds between the concept pairs, or whether no relation holds (\textsc{Random}) (cf. Fig. \ref{fig:pipe}, Step 3 (left) for examples). 

\textbf{Linking Concepts with \textit{Multihop Paths.}}
COMET requires as input a source concept and a relation. For each concept pair $c_s,c_t$ 
we build such inputs by combining $c_s$ with each relation $r_i$ $\in$ $R_{CN13}$, yielding 13 pairs $c_s,r_i$ which we input  for target prediction (Step 2, 
right).
To discover relation chains starting from $S_2$, 
we apply
the same process
to 
$c_t$, using
$c_s$ as target concept. 
We also include \textit{inverse relations}, which gives us greater 
flexibility for connecting entities, i.e., paths are allowed to contain inverted triplets (e.g. \textit{baking $\leftarrow$ \textsc{UsedFor} $\leftarrow$ oven $\rightarrow$ \textsc{AtLocation} $\rightarrow$ kitchen}). To this end, 
we switch the order of concept pairs within a given
relation $r_i$, 
relabel the relation as $r^{-1}_i$, and add the inverted relation pair
to COMET's training set. 

\textbf{
Forward Chaining.} For all pairs in the cross-product $c_s\times c_t$, for each input 
$c_s, r_i$ and 
$c_s, r^{-1}_i$
we generate the 10 most confident concepts $c_{t_i}$ with COMET (beamsize 10) trained on \textsc{CN-13} including inverse triples. 
We continue with all paths where the generated concept $c_{t_i}$ has minimum cosine distance of 0.7 to the respective
target concept $c_t$. 
We generate the next hop by using each $c_{t_i}$ as a new source concept, combine it with each of the 13 original and inverse relations, generate 
novel target predictions, and 
again compare to the 
target concept.
This 
similarity comparison 
guides the forward chaining process towards the
chosen target concepts and helps detecting contextually relevant paths. We use ConceptNet numberbatch embeddings for the encoding of concepts; for multiword concepts we average the embeddings of all non-stopwords.


\textbf{Terminating Paths.} We terminate path generation as soon as 
the similarity between 
$c_{t_i}$ and $c_t$ 
is higher than 0.95 -- here we expect the two concepts to express the same meaning.
We restrict the path length 
to 3 hops and consider only completed paths for evaluation (Step 3, right in Fig. \ref{fig:pipe}).

\textbf{Combining Approaches.}
With our framework 
\textsc{Co-Nnect} we leverage
the potential of
the complementarity of the two model types 
by 
combining COREC-LM and COMET 
in a straightforward way.
Our hypothesis is that 
a system that  
admits both single and multihop connections for establishing links between concepts offers the
greatest flexibility.
We further hypothesize that direct relations 
should be preferred over indirect multihop paths, since the latter
could include irrelevant or misleading intermediate nodes. 
Thus, 
we discard all multihop paths 
for each concept pair for which 
COREC-LM predicted a 
direct connection 
(Fig.\ \ref{fig:pipe}, Step 4, pair 4). 
If COMET \textit{and} COREC-LM 
produce a singlehop path,
we also prefer COREC-LM's prediction, relying on the model's high precision (pair 1 in Step 4). 
We keep the paths generated by COMET for concept pairs 
for which no direct relation could be 
established
(i.e.,
COREC-LM predicted \textsc{Random} or no prediction above its threshold, pair 3\&6), 
assuming 
that in such cases
intermediate concepts are required to establish a 
link. 
If only one of the models establishes a link,  
we keep this connection (pair 2), 
and if none of the models finds a link, 
we assume that the concepts are not (closely) connected (pair 5). 

\subsection{Static Baseline Model}

We compare COREC-LM and COMET against the model of \citet{paul2019ranking} that uses ConceptNet as a static KG. 
The system extracts paths between pairs of concepts from sentence pairs, hence conforms well to 
our setting.
Following \citet{paul2019ranking}, starting from concepts in a sentence pair (\S 3.2), we construct a subgraph $G'$ = $(V',E')$ of the ConceptNet graph, where $V'$ comprises all concepts $c_i$ in $\langle S_1,S_2\rangle$. The system then finds all shortest paths $p'$ from ConceptNet that connect any concept pairs in $V'$, and includes them in $G'$. It then
includes, for any concepts in $G'$,
all directly connected concepts from ConceptNet 
together with their edges. This yields a small sub-graph from ConceptNet that contains concepts and relations relevant for capturing conceptual links  across the sentence pair.   
To select \textit{relevant} paths, $G'$ is filtered by computing scores for vertices and paths using PageRank and Closeness Centrality score, and we constrain path lengths to 3 hops.

\section{Revealing Implicit Knowledge through Knowledge Paths: Experiments and Evaluation}
\label{sec:experiments}

\begin{figure*}
    \centering
    \includegraphics[width=1
\textwidth]{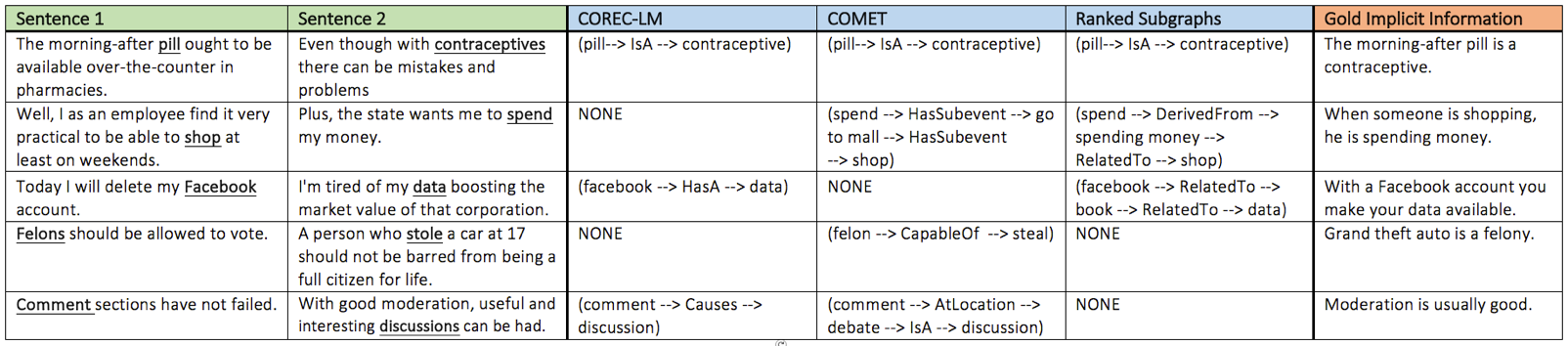}
    \caption{Example generations from our three model types, where the first three instances come from IKAT, and the last two from ARC.}
    \label{fig:exa}
\end{figure*}

In this section we evaluate the paths generated by our proposed models. 
We first present our datasets 
and statistics on established connections (\S 4.1), 
and then evaluate the quality of the paths manually (\S 4.2) and automatically (\S 4.3). 

\subsection{Datasets and Statistics} 


The \textbf{IKAT} dataset \cite{beckeretal:2020a} is based on the English Microtexts Corpus
of short argumentative texts \cite{peld:stede:2015}. For all sentence pairs that are 
adjacent or argumentatively related, annotators added the implicit knowledge that connects
them, using short 
sentences. IKAT contains 719 such sentence pairs, from which we extracted 60,294 concept pairs. 
The \textbf{ARC} dataset \cite{habernal-etal-2018-argument} 
contains arguments taken from online discussions in English, consisting of a claim and a premise, and an annotated implicit warrant that explains why the claim follows from the premise. We evaluate our models on the ARC test set that comprises
444 argument pairs, from which we extracted 21,898 concept pairs; and the corresponding warrants. 

\textbf{Example generations} for both datasets from our three model types -- COREC-LM, COREC, and ranked CN-graphs -- appear in Fig. \ref{fig:exa}, where the first three sentence pairs come from IKAT, and the last two from ARC.

\textbf{Number of links and hops.} Table \ref{tab:stats} gives statistics of the paths generated between concepts from sentence pairs from IKAT and ARC using our  
different models.
We find that COREC-LM finds relations between around 22k from 66k concept pairs in \textbf{IKAT}, while COMET only generates paths between 3,660 pairs. This can be
explained by 
the very high
similarity threshold we imposed for guiding the forward chaining process towards the target concept, since our motivation was not to generate as \textit{many} paths as possible, but paths that are \textit{meaningful} and contextually \textit{relevant}. 
When combining paths from COMET and COREC-LM, 
we find links for more than 24k concept pairs in IKAT. The highest number of
links is
established by ranking CN-subgraphs 
(50k linked concept pairs).
For \textbf{ARC}, which contains 22k concept pairs, COREC-LM finds links between around 10k and COMET around 2k concept pairs, while almost 15k pairs can be connected using ranked CN-graphs. 
In both datasets, the ranked CN-graphs contain on average 2.1 \textbf{hops} (relations) per path,  while the paths generated by COMET are shorter (1.4  
on IKAT/1.5 on ARC). In fact, COMET establishes many direct relations (69\% of all 
paths are single hops), whereas the ranked CN-graphs are mostly two- (49\%) or three-hop paths (37\%).

\textbf{Replacing Vague Relations in CN-Graphs.}  We find that in contrast to COREC-LM and COMET, the ranked CN-graphs  are constructed using mostly the very general relation \textsc{RelatedTo} (71\%/IKAT; 72\%/ARC), 
followed by the likewise vague relation \textsc{HasContext} (8\% in both datasets).\footnote{For details on relation distributions cf. Appendix.}  
For determining the impact of vague relations on path quality, we replace all \textsc{RelatedTo} and \textsc{HasContext} relations in
the ranked CN-graphs with relations predicted by
COREC-LM (trained on CN-13, threshold 0.9). 
For IKAT, we replace 43.4\% of all \textsc{RelatedTo} and 46.2\% of all \textsc{HasContext} instances, 
in ARC we 
replace 70.7\% of all \textsc{RelatedTo} and 37\% of all \textsc{HasContext} relations. We use this version 
when evaluating paths, in addition to the original ranked CN-graphs.


\begin{table}[]
\centering
\scalebox{0.9}{
\small{
%

\begin{tabular}{l|l|l|l|l|l|l|l|l|l|l}
\toprule
 &  & \textbf{COR} & \textbf{COM} & \textbf{CONN} & \textbf{CN} \\
\toprule
\textbf{IKAT} & linked pairs & 21,934 & 3,660 & 24,063 & 50,003 \\
\cline{2-6}
 & avg. hops & 1 & 1.4 & 1.1 & 2.1 \\
 \cline{2-6}
\toprule
\textbf{ARC} & linked pairs & 9,844 & 1,826 & 10,828 & 14,940 \\
\cline{2-6}
 & avg. hops & 1 & 1.5 & 1.1 & 2.1 \\
   \bottomrule
\end{tabular}}}
\caption{Statistics of paths generated by COREC-LM, COMET, their combination (\textsc{Co-Nnect}), and ranking CN-graphs (CN): number of concepts pairs between which a link was found, and average number of hops per path. 
}
\label{tab:stats}
\end{table}

\subsection{Manual Evaluation of Path Quality} 

Our statistics showed that most links between concepts can be revealed using knowledge paths retrieved from ConceptNet as a static KG, whereby these paths tend to con\-tain multiple hops and a high amount of vague relations. Fewer links are es\-tab\-lished using the dynamic models COREC-LM and COMET, which produce shorter paths using only specific relation types from CN-13. Since our aim is to construct high-quality, meaningful knowledge paths that help to explain implicit information (rather than establishing as many links as possible), we now examine
the quality and relevance of the 
knowledge
paths. 
We set up an annotation experiment, 
providing annotators with 100 sentence pairs 
from each dataset, with marked concepts (one from $S_1$ and one from $S_2$) 
and the 
path  
generated between these concepts by (i) COREC-LM, (ii) COMET, (iii) ranked paths from CN, and (iv) ranked paths 
with replaced vague relations (CN-r). 

\textbf{Annotation Setup.} For each sentence pair, our annotators evaluated if 1) the path is a meaningful and relevant explanation for the connection between the two sentences (very relevant/relevant/neutral/not relevant/misleading); if 2) the path represents implicit information not explicitly expressed in the sentences (yes/no); and 3) which model 
generates the path that is most helpful and expressive for understanding the connection between the sentences. 
4) To evaluate the \textit{combination} of COREC-LM and COMET in \textsc{Co-Nnect}, we generate a subset for each dataset that includes all sentence/concept pairs for which COREC-LM predicted a 
singlehop path \textit{and} COMET generated a multihop path (10 pairs per subset). For these instances we ask in addition whether the multihop paths include unrelated, unnecessary or uninformative intermediate nodes (yes/no), misleading intermediate nodes (yes/no); or intermediate nodes that are important for explaining the connection and missing in the direct relation predicted by COREC-LM (yes/no).\footnote{The annotation manual together with example annotations can be found here: \url{https://github.com/Heidelberg-NLP/CO-NNECT/blob/main/manual.pdf}}
Annotations were performed by two annotators with a linguistic background. We measure IAA using Cohen’s Kappa and achieve an agreement of {81\%}. Remaining conflicts were resolved by an expert annotator. 

\begin{table}[t!]
    
\scalebox{0.7}{
  \centering
\small{
\begin{tabular}{cc|cccc|cccc}
\toprule
&&\multicolumn{4}{c|}{\textbf{IKAT}}&\multicolumn{4}{c}{\textbf{ARC}}\\
 \toprule
 &  & \small{COR} & \small{COM} & \small{CN} & \small{CN-r} & \small{COR} & \small{COM} & \small{CN} & \small{CN-r} \\
\toprule
Predictions &  & 74 & 64 & \textbf{88} & \textbf{88} & \textbf{78} & 60 & 76 & 76\\
 \hline
Relevance 
& +2 & \textbf{70} & 50 & 36 & 40 & \textbf{63} & 49 & 30 & 34\\
 & +1 & 19 & \textbf{27} & 22 & 24 & 25 & 28 & 28 & \textbf{32}\\
 & 0 & 8 & 18 & 27 & 21 & 8 & 9 & 29 & 22\\
 & -1 & 3 & 5 & 10 & 10 & 2 & 6 & 4 &3\\
 & -2 & 0 & 0 & 5 & 5& 2 & 8 & 9 & 9\\
  \hline
Implicit  
 & yes & \textbf{80} & 78 & 57 & 67 &\textbf{87} & 81 & 57  & 62\\
Knowledge & no & 20 & 22 & 43 & 33 & 13 & 19 & 43 & 38\\
  \hline
Best link &  & \textbf{65} & 64 & 28 & 34 &\textbf{76} & 70 & 7 & 14\\
 \bottomrule
\end{tabular}}}
\caption{\small{Manual evaluation of paths
from COREC-LM, COMET, ranked CN-graphs (CN), and CN-graphs with replaced vague relations (CN-r); all numbers in \%.}} 
\label{tab:manualanno}
\end{table}

\textbf{Results.} Table \ref{tab:manualanno} shows the results of our annotation experiment. 
On \textbf{IKAT}, 89\% of the paths established by COREC-LM and 77\% of the relations predicted by COMET were annotated as very relevant (+2) or relevant connections (+1), 
which only applies for 58\% of the ranked CN-paths. 
15\% of the ranked CN-paths were annotated as not relevant (-1) or 
misleading (-2), which can be explained by noisy intermediate nodes; 
and 27\% as vague (0), which can be explained by the large amount of 
unspecific relations. 
When replacing \textsc{RelatedTo} and  \textsc{HasContext} (CN-r), the amount of paths annotated as vague slightly decreases, and the amount of paths labelled as relevant and very relevant increases.

Moreover, paths generated by COREC-LM and COMET were found to yield better implicit knowledge representations than ranked CN-paths
(line 8-9, Table \ref{tab:manualanno}), 
while we find that replacing vague relations in the CN-paths improves their ability of  representing  implicit knowledge. 
Finally, 65\% of  relations predicted by COREC-LM and 64\% of paths generated by COMET were chosen as explaining the connections between sentences best, which is only the case for 28\% of the CN-paths, and slightly better for the replaced version of the CN-paths (34\%).

On \textbf{ARC}, 
the high amount of CN-paths annotated as vague (29\%) again indicates 
uninformative connections and can be reduced when replacing 
vague with more specific relations. 
Relations predicted by COREC-LM were found to be less relevant for connecting sentences in ARC than in IKAT, but 87\% of them were evaluated as appropriate expressions of implicit knowledge. 
76\% of the relations predicted by COREC-LM were evaluated as best connections, which applies only for 7\% of CN-paths and 
14\% of CN-paths with replaced 
relations. 
For COMET we find overall comparable results between IKAT and ARC. 

Regarding the \textbf{combination} of COREC-LM and COMET addressed with question 4,
according to our annotators 50\% of the multihop paths in the IKAT subset include misleading nodes and \textit{all} of them include irrelevant or uninformative nodes. Still, compared against the direct relations predicted by COREC-LM, annotators state for 30\% of the multihop paths from COMET that they contain intermediate concepts that are important for explaining the connection.  
On the ARC subset, 40\% of the multihop paths include misleading and 60\% include irrelevant nodes, and only 20\% contain important intermediate concepts that are missing in the direct relation. For each subset, annotators preferred the shorter path over the multihop path in 90\% of the given sentence pairs. 
Comparing singlehop paths generated by COMET to direct relations predicted by COREC-LM for the same concept pairs, 
our annotators preferred the relation predicted by COREC-LM in 64\% of the cases, in 29\% the link was annotated as equally good, and only in 7\% COMET's generation was preferred. 

To summarize, according to our manual evaluation,
the dynamic models COMET and COREC-LM 
are better suited for generating meaningful knowledge paths that express
implicit knowledge between sentences than ranked paths from the static CN
knowledge graph,
even though
replacing 
vague 
by more specific relations
slightly improves results. 
When comparing multihop paths to direct relations established between the same concept pairs, we find that longer paths tend to contain irrelevant or even misleading nodes, and that direct relations are preferred by human annotators. 
These findings support our proposed joint framework \textsc{Co-Nnect},  which 
gives preference to
direct relations and utilizes multihop paths only if no direct connection between concepts can be revealed.

\subsection{Automatic Evaluation Against Gold} 

Our goal is to generate meaningful 
paths that 
convey
implicit knowledge 
between 
sentences. In our automatic evaluation
we 
compare the set of model-generated paths 
between all concept pairs from two related sentences to the implicit knowledge annotated in IKAT 
and ARC 
for these
sentences, using 
similarity metrics.

Since the generated relation and path representations 
differ
from the annotated natural sentences, we approximate a common representation as follows:
We \textbf{encode the golden annotations of implicit knowledge} -- 
usually short 
sentences -- 
using
three 
settings: 
(i) \textbf{Silver Paths:} we encode 
their 
relational knowledge, by extracting
all concepts from
each golden implicit knowledge sentence
(\textit{My dog has a bone} $\rightarrow$ \textit{dog, bone}) using the CN-extraction tool \textsc{CoCo-Ex} \cite{beckeretal:2021:COCOEX}, and 
predict the relations between them using COREC-LM, trained on CN-13 (\textit{dog}, \textsc{HasA}, \textit{bone}). If a sentence contains more than two relations, we concatenate the predicted relation triples. 
(ii) IKAT provides manual annotations of ConceptNet relations for the golden implicit knowledge sentences, which we use as \textbf{Gold Paths} (\textit{The tree is in the garden} $\rightarrow$ \textit{tree} \textsc{AtLocation} \textit{garden}). 
(iii) \textbf{Gold-NL}: Here we use the 
implicit knowledge (in natural language) 
as provided in the datasets: IKAT's 
implicit knowledge sentences 
and ARC's
implicit warrants.

For \textbf{encoding the generated paths} 
we apply two settings: (i) we concatenate all concepts and relations within the paths; (\textbf{Generated Paths}) and (ii) we translate the relation triples and paths to (pseudo) natural language using templates provided by ConceptNet (e.g. $c_s$ \textsc{Causes} $c_t$ $\rightarrow$ \textit{The effect of} $c_s$ \textit{is}  $c_t$; 
\textbf{Generated Paths-NL}). 

We 
apply two \textbf{automatic similarity metrics}, comparing (a) Generated 
vs.\ Silver Paths, (b) Generated Paths-NL vs.\ Gold-NL, and (c) Generated 
vs.\ Gold Paths (only IKAT). 
(i) We encode each 
representation as described above using ConceptNet numberbatch embeddings \cite{speer17} (for multiword concepts we average the embeddings of all non-stopwords), and compute cosine similarity between them,
and (ii) we use
\textsc{BertS}core F1 \cite{bert-score} to compare
representations, 
which computes 
string similarity 
using contextualized embeddings. Both metrics lie in
[$-$1, 1].

\textbf{Results.} Table \ref{tab:eval-sim} shows that the paths generated by combining COREC-LM and COMET in our framework \textsc{Co-Nnect} achieve the highest similarity scores according to Numberbatch-Cosim on \textbf{IKAT} in setting (a) and (b), while for (c) we get the highest Cosim scores for ranked CN-graphs with replaced vague relations. According to \textsc{Bert}Score, either COREC-LM (setting a) or COMET (setting b) applied separately, or both applied in combination (setting c) achieve highest results on IKAT. On \textbf{ARC}, \textsc{Co-Nnect} achieves both highest Cosim and \textsc{Bert}Scores in setting (a), while in (b) we get the best scores for CN-r according to Cosim, and the best scores for COMET according to \textsc{Bert}Score.

\begin{table}[t!]
      \centering
        \scalebox{1.0}{
\small{
\begin{tabular}{@{}l|lllll@{}}
 \toprule
  & \textbf{COR} & \textbf{COM} & \textbf{CONN} & \textbf{CN} & \textbf{CN-r}\\
 \toprule
 \multicolumn{6}{c}{\textbf{(a) Generated Paths vs. Silver Paths}} \\
 IKAT  &  .61/\textbf{.85} & .54/.82 & \textbf{.62}/.84  & .57/.78 & .58/.80 \\
  ARC  &  .41/.84 & .39/.82&  \textbf{.42}/\textbf{.86} & .40/.77 & .40/.78 \\
\toprule
\multicolumn{6}{c}{\textbf{(b) Generated Paths-NL vs. Gold-NL}}  \\
IKAT   & .69/.81 & .65/\textbf{.83} & \textbf{.70}/.81 & .65/.75 & .69/.76 \\
ARC  & .72/.81 & .66/\textbf{.82} & .72/.81 & .71/.75 & \textbf{.77}/.76 \\
\toprule
\multicolumn{6}{c}{\textbf{(c) Generated Paths vs. Gold Paths}} \\
IKAT &.57/.78 & .49/.78 & .58/\textbf{.79} & .66/.73 & \textbf{.67}/.74 \\

  \bottomrule
\end{tabular}}}
\caption{\small{Comparing generated paths to implicit knowledge annotations on IKAT and ARC, measured by Cosim/\textsc{Bert}Score (F1).}}
\label{tab:eval-sim}
\end{table}

Summarizing our insights from  automatic evaluations, 
we find that \textbf{COREC-LM} achieves high scores when applied separately \textit{or} in combination with COMET (\textsc{Co-Nnect}). \textbf{COMET} applied in isolation
does not yield the 
highest 
scores, but 
helps to boost COREC-LM's performance in the 
joint \textsc{Co-Nnect} framework. 
\textbf{Ranked CN-graphs} 
achieve highest Cosim 
in two settings/datasets (ARC--b; IKAT--c), but
we do not find significant improvements 
when replacing vague relations in CN-graphs (expect for Cosim in 
setting b). This can be explained by the fact that 
even though many 
\textsc{RelatedTo} and \textsc{HasContext} instances
could be replaced, for both datasets a large amount of vague relations still remain 
(56.6\% of  \textsc{RelatedTo}/53.2\% of \textsc{HasContext} in IKAT; 29.3\% \textsc{RelatedTo}/63\% \textsc{HasContext} in ARC). Therefore, the vague relation types in the 
CN-graphs still remain problematic when representing implicit knowledge. 

When comparing 
our 
\textbf{manual}  evaluation results 
to the \textbf{automatic} 
scores, 
we find that the generations 
that were manually evaluated as most relevant and meaningful explanations of implicit knowledge are not always 
highest-ranked by automatic
metrics, which
points to two 
limitations of our automatic evaluation: Besides 
well-known issues regarding the reliability, interpretability, and biases of automatic 
metrics \cite{callison-burch-etal-2006-evaluating}, we evaluate
the generated paths against an 
annotated \textit{reference} -- 
paths or sentences -- which is often
only one among several valid options for expressing the 
implicit knowledge 
(cf.\ \citeauthor{beckeretal:2017a} \citeyear{beckeretal:2017a}). This means that a generated path may
still be a relevant explicitation of
implicit information, even if 
\textit{not}  similar to the reference.
Hence, automatic scores are 
to be considered with caution.

\section{Conclusion}
\label{sec:conclusion}

Our work aims to
leverage commonsense knowledge in the form of 
single and multihop paths, to establish knowledge connections between concepts from different 
sentences, as a form of explicitation of implicitly conveyed information.
We combine existing relation classification 
and target prediction 
models in a dynamic knowledge prediction framework, \textsc{Co-Nnect}, utilizing language models finetuned
on knowledge relations from ConceptNet.
We compare against 
a 
path ranking system that employs static knowledge from ConceptNet as a baseline and
evaluate the quality of the obtained paths (i) through manual evaluation
and (ii) using automatic similarity metrics, by comparing generated paths to annotations of implicit knowledge 
in two argumentative datasets.
Our 
evaluations show that 
we obtain 
the highest number of connections 
from the static ConceptNet graph, however, they
are often noisy due to 
unrelated intermediate nodes, and -- even after replacements -- still contain many 
unspecific relations.
Our framework \textsc{Co-Nnect}, instead,
combines  relation classification and target prediction, leveraging the \textit{high precision} of the former, and the \textit{ability to perform forward chaining} of the latter, and obtain 
high-quality, meaningful and relevant knowledge paths that reveal implicit knowledge conveyed by the text, as shown in a profound manual evaluation experiment.

We believe that \textsc{Co-Nnect} is a useful framework which can be applied for different tasks, such as enriching
texts with commonsense knowledge relations and paths, for dynamically enriching
knowledge bases, or for building knowledge constraints for language generation.
In \citet{beckeretal:2021:lm} for example we inject single- and multi-hop commonsense knowledge paths predicted by \textsc{Co-Nnect}
as constraints into language models and show 
that this improves the model's ability of generating sentences that explicate
implicit knowledge which connects sentences in texts.
We furthermore believe that the paths established with \textsc{Co-Nnect}, which can provide explicitations of implicit  knowledge, 
can be useful to enhance many other
NLP downstream tasks, such as argument classification, stance detection, or commonsense reasoning. 

\section*{Acknowledgements}
This work has been funded by the DFG within the project
ExpLAIN as part of the Priority Program “Robust Argumentation Machines” (SPP-1999). We thank our annotators for their contribution.




\bibliographystyle{acl_natbib}
\bibliography{anthology,acl2021}

\section*{APPENDIX}
\begin{table*}[!t]
\centering
\small{
\begin{tabular}{l|l|l|l|l}
 &  \textbf{COREC-LM} & \textbf{COMET} & \textbf{CONNECT} & \textbf{CN Subgraphs} \\
\hline \hline
 \textbf{IKAT} & \textsc{AtLocation}(25\%) & \textsc{IsA}(19\%) & \textsc{AtLocation}(22\%) & \textsc{RelatedTo}(71\%) \\
 &  \textsc{HasProperty}(20\%) & \textsc{HasA}(18\%) & \textsc{IsA}(17\%) & \textsc{HasContext}(8\%) \\
 &  \textsc{IsA}(17\%) & \textsc{Causes}(17\%) & \textsc{HasProperty}(16\%) & IsA(7\%) \\
\hline
\textbf{ARC} & \textsc{AtLocation}(31\%) & \textsc{AtLocation}(22\%) & \textsc{AtLocation}(27\%) & \textsc{RelatedTo}(72\%) \\
 & \textsc{IsA}(18\%) & \textsc{Causes}(20\%) & \textsc{IsA}(15\%) & \textsc{HasContext}(8\%) \\
 & \textsc{HasProperty}(14\%) & \textsc{HasA}(18\%) & \textsc{HasProperty}(10\%) & \textsc{IsA}(7\%)\\
\end{tabular}}
\caption{Most frequently used relations when constructing single and multihop knowledge paths using COMET, COREC-LM, their combination, and ranked subgraphs from CN.}
\label{tab:rels}
\end{table*}

\subsection*{A Constructing the \textsc{Random} Class for Training COREC-LM in an Open World Setting} Our downstream application task -- finding connections between concepts -- requires that our relation classifier also learns to detect that no direct relation holds between a given pair of concepts. We thus
extend the data for training and testing COREC-LM with a \textsc{Random} class which contains concept pairs that are not
related which we add to CN-100k and CN-13 Instances for this class are generated similarly to \citet{vylomova-etal-2016-take}: 
50\% of them are opposite pairs which we obtain by switching the order of concept pairs within the same relation, and 50\% are corrupt pairs, obtained by replacing one concept in a pair with a random concept from the same relation. Corrupt pairs ensure that COREC-LM learns to encode relation instances rather than simply learning properties of the word classes. We add these instances (2070 for training and 260 for development and testing, respectively) to CN-100k and CN-13 when training and evaluating in an open world setting.


\subsection*{B Relations Used for Constructing Single- and Multihop Paths} Table \ref{tab:rels} lists the three most frequently used relations when constructing single and multihop knowledge paths using COMET, COREC-LM, their combination, and ranked subgraphs, respectively for the two datasets IKAT and ARC. The top three relations used by \textbf{COREC-LM} within both datasets are \textsc{AtLocation}, \textsc{HasProperty}, and \textsc{IsA}. Interestingly, besides \textsc{IsA} and \textsc{HasA}, \textbf{COMET} frequently uses the only causal relation in the CN inventory  \textsc{Causes}.  In contrast to COREC-LM and COMET, the ranked CN-graphs  are constructed using mostly the very general relation \textsc{RelatedTo}, followed by the likewise vague relation \textsc{HasContext}. When excluding paths that contain \textsc{RelatedTo}, only 2,551 connected concept pairs remain in IKAT and 6,858 in ARC.

\end{document}